\documentclass[lettersize,journal]{IEEEtran}

\usepackage{graphicx} 
\usepackage{epstopdf}
\usepackage{subfigure}
\usepackage{stfloats}
\usepackage{algorithm, algorithmic}
\usepackage{diagbox}
\usepackage{subcaption} 
\usepackage{caption} 
\usepackage{multirow}
\usepackage{mathtools}
\usepackage{setspace}  
\usepackage{threeparttable}
\usepackage{textcomp,booktabs}
\usepackage[usenames,dvipsnames]{color}
\usepackage{colortbl}
\usepackage{indentfirst}
\usepackage{cite}
\usepackage{amsmath,amssymb,amsfonts}
\usepackage{algorithmic}
\usepackage{graphicx}
\usepackage{textcomp}
\usepackage{xcolor}
\usepackage{mathtools}
\usepackage{amsthm}
\usepackage{color}
\usepackage{enumitem}
\theoremstyle{definition}

\definecolor{mygray}{gray}{.9}
\definecolor{mypink}{rgb}{.99,.91,.95}
\definecolor{mycyan}{cmyk}{.3,0,0,0}

\begin{document}
\title{Mixture of Experts-augmented Deep Unfolding for Activity Detection in IRS-aided Systems}
\author{Zeyi Ren, Qingfeng Lin, Jingreng Lei, Yang Li, and Yik-Chung Wu 
\thanks{The work of Y. Li was supported in part by Guangdong Basic and Applied Basic Research Foundation under Grant 2025A1515011658, and in
part by the National Natural
Science Foundation of China (NSFC) under Grant 62101349. (Corresponding authors: Yang Li and Yik-Chung Wu.)}
\thanks{Z. Ren, Q. Lin, and Y.-C. Wu are with the Department of Electrical and Electronic Engineering, The University of Hong Kong, Hong Kong (e-mail: \{renzeyi, qflin, ycwu\}@eee.hku.hk).}
\thanks{J. Lei is with the Department of Electrical and Electronic Engineering, The University of Hong Kong, Hong
Kong, and also with the School of Computing and Information Technology, Great Bay University, Dongguan 523000, China (e-mail: leijr@eee.hku.hk).}
\thanks{Y. Li is with the School of Computing and Information Technology, Great Bay University, Dongguan 523000, China, and also with Dongguan Key Laboratory for Intelligence and Information Technology, Dongguan 523000, China (e-mail: liyang@gbu.edu.cn).}
}

\maketitle

\begin{abstract}
In the realm of activity detection for massive machine-type communications, intelligent reflecting surfaces (IRS) have shown significant potential in enhancing coverage for devices lacking direct connections to the base station (BS). However, traditional activity detection methods are typically designed for a single type of channel model, which does not reflect the complexities of real-world scenarios, particularly in systems incorporating IRS. To address this challenge, this paper introduces a novel approach that combines model-driven deep unfolding with a mixture of experts (MoE) framework. By automatically selecting one of three expert designs and applying it to the unfolded projected gradient method, our approach eliminates the need for prior knowledge of channel types between devices and the BS. Simulation results demonstrate that the proposed MoE-augmented deep unfolding method surpasses the traditional covariance-based method and black-box neural network design, delivering superior detection performance under mixed channel fading conditions.
\end{abstract}

\vspace{-2mm}
\section{Introduction} \label{intro}
\vspace{-2mm}
Massive machine-type communications (mMTC) have been expected to play a vital role to empower the sixth-generation (6G) vision of future ubiquitous connectivity~\cite{isacqf}. To meet the low latency requirement in mMTC, grant-free random access is recognized as a promising solution. However, grant-free random access requires the base station (BS) to perform activity detection~\cite{Liuliang_TSP_FUTURE,tsp_hao,liyang_ad}. Due to the large number of potential Internet of Things (IoT) devices and nonorthogonal signature sequences, activity detection is a challenging task.

Mathematically, the optimization algorithms for activity detection task have been broadly categorized into two types: compressed sensing (CS)-based algorithms~\cite{tspll} and covariance-based algorithms~\cite{haghighatshoar2018, C_ICC_Czl_ad, TWC_Lin2022}. Both theoretically and empirically,
it has been demonstrated that the covariance-based algorithms
generally outperform the CS-based algorithms in terms of detection performance~\cite{tit_c}. However, covariance-based algorithms rely on the tractability of the covariance matrix of the received signal, which further requires accurate channel fading statistics from various devices to the BS. This might not be practical, especially in the recent intelligent reflecting surface (IRS)-aided systems~\cite{J_TWC_shaoxiaodan_risad,J_TSP_shiyuanming_risad} where some devices may directly connected to the BS while other through the help of an IRS. 

On the other hand, deep learning-based algorithms have become popular in communication research due to their ability to overcome modeling inaccuracy~\cite{Sun_TSP,lei}. However, black box deep learning designs (e.g., multi-layer perceptrons (MLP) or convolutional neural networks) do not incorporate the domain knowledge of communication systems. This usually results in unsatisfactory performance or the neural network not being able to converge during training. To overcome such drawbacks, model-driven deep unfolding has emerged as a viable alternative. By regarding each iteration of an optimization algorithm as one layer of the neural networks, the model-driven deep unfolding algorithm embeds the domain knowledge into neural network designs while leveraging on the data to determine the behavior of the model.

From the above arguments, applying deep unfolding to the activity detection problem in IRS-aided systems seems to be an obvious choice. However, there are still two challenges. Firstly, due to the existence of the IRS composite channel, the rank-one update usually employed in covariance-based method is not applicable anymore.  To this end, we propose a projected gradient descent update to facilitate the unfolding. Secondly, even after unfolding, the algorithm still depends on the knowledge of each device's fading type. To circumvent this issue, we leverage the mixture of experts (MoE) approach~\cite{Du_MoE,wc_du}, which takes in the received signal and determines which type of fading channels dominates so that an appropriate expert can be used. In this way, there is no need to know the fading channel type for each device to execute the algorithm.

Numerical results demonstrate that the proposed MoE-augmented model-driven deep unfolding outperforms conventional covariance-based methods and black-box neural network designs. Furthermore, the performance loss from not knowing the fading channel type of each device is minimal compared to the unfolded network with perfect information.

\section{System Model and Problem Formulation}   \label{sys}

\begin{figure} [t]
	\centering 
		 \includegraphics[width=0.4\textwidth]{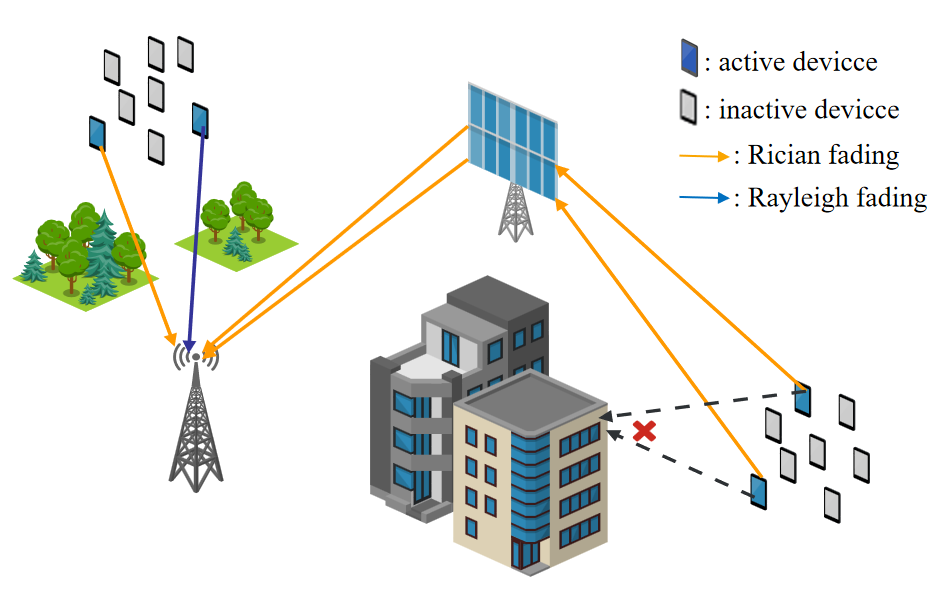} 
  	\caption{IRS-aided massive random access with mixed channel types}\vspace{-0.2in}  \label{Fig7}
\end{figure}

Consider a single-cell network with a BS having $M$ antennas, as shown in Fig.~\ref{Fig7}. In this network, there are three sets of devices, with the sets denoted by $\mathcal{K}_1$, $\mathcal{K}_2$, and $\mathcal{K}_3$, respectively. For devices in $\mathcal{K}_1$, their direct links to the BS are obstructed, thus require an IRS with $N$ reflecting elements to enhance communication quality~\cite{J_TWC_shaoxiaodan_risad,J_TSP_shiyuanming_risad}. For $\mathcal{K}_2$ and $\mathcal{K}_3$, the devices are those only with direct links to the BS, with devices in $\mathcal{K}_2$ following Rician fading and those in $\mathcal{K}_3$ following Rayleigh fading.

Let $\mathbf{f}_k \in \mathbb{C}^{M}$ denote the small-scale fading from device~$k$ to the BS, it can be expressed as
\begin{align}\label{f}
    \mathbf{f}_k \sim \sqrt{\frac{\kappa_{\text{U}}}{1+\kappa_{\text{U}}}} \mathbf{f}_k^{\text{LoS}}+\sqrt{\frac{1}{1+\kappa_{\text{U}}}} \mathcal{CN} \left(\mathbf{0}, \mathbf{I}_{M}\right),
\end{align}
where $\mathbf{f}_k^{\text{LoS}}$ is given by the array response vector at the BS due to device~$k$, and $\kappa_{\text{U}}$ is the Rician factor. In case of Rayleigh fading, $\kappa_{\text{U}}$ = 0. Similarly, the small-scale fading from device~$k$ to the IRS and that from the IRS to the $m$-th antenna of the BS are modeled as:
\begin{align}\label{h} 
\mathbf{h}_k \sim \sqrt{\frac{\kappa_{\text{R}}}{1+\kappa_{\text{R}}}} \mathbf{h}_k^{\text{LoS}}+\sqrt{\frac{1}{1+\kappa_{\text{R}}}} \mathcal{CN} \left(\mathbf{0}, \mathbf{I}_{N}\right), 
\end{align}
\begin{align} \label{g}
\mathbf{g}_m \sim \sqrt{\frac{\kappa_{\text{B}}}{1+\kappa_{\text{B}}}} \mathbf{g}_m^{\text{LoS}}+\sqrt{\frac{1}{1+\kappa_{\text{B}}}} \mathcal{CN} \left(\mathbf{0}, \mathbf{I}_{N}\right), 
\end{align}
where $\kappa_{\text{R}}$ and $\kappa_{\text{B}}$ are the corresponding Rician factors.

Denote the device activity indicator of device~$k$ as $b_k = 1$ if it is active and 0 otherwise.
Moreover, a unique signature sequence ${\mathbf{s}}_{k}=\left[ s_{k, 1}, s_{k, 2}, \ldots, s_{k, L}\right]^{T} \in \mathbb{C}^{L}$ is assigned to each device~$k$. At the start of each block, all the active devices send their signature sequences and the BS conducts activity detection based on the received signals. Assuming that the transmissions from different devices are synchronous, the received signal is
\begin{align}\label{receive}
\begin{aligned}
\mathbf{Y} &= \sum_{k\in\mathcal{K}_1} \underbrace{b_{k}  \sqrt{p_k \beta_{k}}}_{a_k}\mathbf{s}_{k}   {\mathbf{h}}^{H}_{k} \mathbf{\Theta} \mathbf{G} + \sum_{k\in \mathcal{K}_2} b_{k}\sqrt{p_k \beta_{k}}\mathbf{s}_{k}   {\mathbf{f}}^{H}_{k}\\
&+ \sum_{k\in \mathcal{K}_3} b_{k}\sqrt{p_k \beta_{k}}\mathbf{s}_{k}   {\mathbf{f}}^{H}_{k} +\mathbf{W},
\end{aligned}
\end{align}
where $\mathbf{G}=[\mathbf{g}_1,\mathbf{g}_2,\dots,\mathbf{g}_M]$,  $\beta_{k}$ is the cascaded large-scale fading coefficient for device $k$, $p_{k}$ is the transmit power of the $k$-th device, $\mathbf{\Theta}$ is a diagonal matrix containing the phase-shift coefficients of the IRS elements. The elements of $\mathbf{W} \in \mathbb{C}^{L\times M}$ are independent and identically distributed (i.i.d.) Gaussian noise at the BS following $\mathcal{C}\mathcal{N}\left(0, \sigma_{w}^{2}\right)$ with $\sigma_{w}^{2}$ being the noise power.

The activity detection problem is mathematically equivalent to detecting whether $a_{k}\triangleq b_{k}  \sqrt{p_k\beta_k}$ is positive or zero. To this end, we treat the device activities $\{a_k\}_{k\in\mathcal{K}_1\cup\mathcal{K}_2\cup\mathcal{K}_3}$ as a set of unknown but deterministic parameters, and model the received signal $\mathbf{Y}$ as a random variable. In particular, denoting the the non-line-of-sight parts of~\eqref{h} and~\eqref{g} as $\mathbf{h}_k^{\text{NLoS}}\sim   \mathcal{CN} \left(\mathbf{0}, \mathbf{I}_{N}\right)$ and $\mathbf{g}_m^{\text{NLoS}}\sim   \mathcal{CN} \left(\mathbf{0}, \mathbf{I}_{N}\right)$. Based on~\cite[Proposition 1]{twc_LIN}, we have~$(\mathbf{h}_k^{\text{NLoS}})^H \mathbf{\Theta} \mathbf{g}_m^{\text{NLoS}} \sim   \mathcal{CN} \left(\mathbf{0}, N\mathbf{I}_{M}\right)$ when $N$ is sufficiently large. Therefore, $\mathbf{Y}$ is a complex Gaussian distributed matrix. The mean of the $m$-th column of $\mathbf{Y}$ (denoted as $\mathbf{y}_m$) is then given by
\begin{align}\label{r1_m}
\bar{\mathbf{y}}_m\triangleq\mathbb{E}\left[ \mathbf{y}_m\right] &=\sum_{k\in\mathcal{K}_1} {a_k}\mathbf{s}_{k}(\sqrt{\frac{\kappa_{\text{R}}}{1+\kappa_{\text{R}}}} \mathbf{h}_k^{\text{LoS}})^H \mathbf{\Theta} \sqrt{\frac{\kappa_{\text{B}}}{1+\kappa_{\text{B}}}} \mathbf{g}_m^{\text{LoS}}\nonumber\\ &+\sum_{k\in\mathcal{K}_2}{a_k}\mathbf{s}_{k}   \sqrt{\frac{\kappa_{\text{U}}}{1+\kappa_{\text{U}}}} \mathbf{f}_{k}^{\text{LoS}}(m)^*,
\end{align}
where $\mathbf{f}_{k}^{\text{LoS}}(m)$ denotes the $m$-th element of $\mathbf{f}_{k}^{\text{LoS}}$. Furthermore, the covariance of $\mathbf{y}_m$ can be computed as
\begin{align} \label{sigma}
&\boldsymbol{\Sigma}_m \triangleq\mathbb{E}\left[ (\mathbf{y}_m - \mathbb{E}\left[ \mathbf{y}_m\right])(\mathbf{y}_m - \mathbb{E}\left[ \mathbf{y}_m\right])^H\right]  \nonumber \\
&=\sum_{k\in\mathcal{K}_1} {a_k}\frac{\kappa_{\text{U}}\|{\mathbf{h}}_k^{\text{LoS}}\|^2}{(1+\kappa_{\text{B}})(1+\kappa_{\text{U}})}\mathbf{s}_{k} \mathbf{s}_{k}^H +\!\! \sum_{k\in\mathcal{K}_1} \frac{{a_k}N\mathbf{s}_{k} \mathbf{s}_{k}^H}{(1+\kappa_{\text{B}})(1+\kappa_{\text{U}})} \nonumber\\
&+ \sum_{k\in\mathcal{K}_1} {a_k}\frac{\kappa_{\text{B}}\|{\mathbf{g}_m}^{\text{LoS}}\|^2}{(1+\kappa_{\text{B}})(1+\kappa_{\text{U}})}\mathbf{s}_{k} \mathbf{s}_{k}^H
+\sum_{k\in\mathcal{K}_2} \frac{a_k\mathbf{s}_{k} \mathbf{s}_{k}^H}{1+\kappa_{\text{U}}} \nonumber\\
&+ \sum_{k\in\mathcal{K}_3} {a_k}  \mathbf{s}_{k} \mathbf{s}_{k}^H +\sigma_w^2 \mathbf{I}_M, \nonumber\\
&=\!\sum_{k\in\mathcal{K}_1} {a_k} \Xi_{m,k} \mathbf{s}_{k} \mathbf{s}_{k}^H \!+\!\!\!\sum_{k\in\mathcal{K}_2} \frac{a_k\mathbf{s}_{k} \mathbf{s}_{k}^H}{1+\kappa_{\text{U}}}+\!\!\!\sum_{k\in\mathcal{K}_3} {a_k}  \mathbf{s}_{k} \mathbf{s}_{k}^H +\sigma_w^2 \mathbf{I}_M, 
\end{align}
where $\Xi_{m,k} =\frac{N +\kappa_{\text{B}}\|{\mathbf{g}}_m^{\text{LoS}}\|^2+\kappa_{\text{U}}\|{\mathbf{h}}_k^{\text{LoS}}\|^2}{(1+\kappa_{\text{B}})(1+\kappa_{\text{U}})}$. Once obtaining the mean and covariance of $\mathbf{y}_m$,
the likelihood function of $\left\{a_k\right\}_{k\in\mathcal{K}_1\cup\mathcal{K}_2\cup\mathcal{K}_3}$ is given by
\begin{align} 
\label{likelihood_z}
&\log p\left(\mathbf{y}_m ; \left\{a_k\right\}_{k\in\mathcal{K}_1\cup\mathcal{K}_2\cup\mathcal{K}_3}\right)=- L \log \pi \nonumber \\& -\left(\log \left|\boldsymbol{\Sigma}_m\right|+\operatorname{tr}\left(\boldsymbol{\Sigma}_m^{-1}\left(\mathbf{y}_m-\bar{\mathbf{y}}_m\right)\left(\mathbf{y}_m-\bar{\mathbf{y}}_m\right)^H\right)\right),
\end{align}
where $L$ denotes the length of signature sequence.

Due to the presence of the IRS, different columns of $\mathbf{Y}$ are not mutually independent. Therefore, an explicit expression for the joint likelihood function of all columns of $\mathbf{Y}$ cannot be obtained. To this end, the activity detection problem can be formulated by minimizing the following approximated negative log-likelihood function $-\log p\left(\mathbf{Y} ;\left\{a_k\right\}_{k\in\mathcal{K}_1\cup\mathcal{K}_2\cup\mathcal{K}_3}\right)\approx \sum_{m=1}^M-\log p\left(\mathbf{y}_m ; \left\{a_k\right\}_{k\in\mathcal{K}_1\cup\mathcal{K}_2\cup\mathcal{K}_3}\right)$, subject to the constraints on $\{{a}_{k}\}_{k\in\mathcal{K}_1\cup\mathcal{K}_2\cup\mathcal{K}_3}$, i.e., 
\begin{subequations} \label{p}
\begin{align} 
& \min _{\{a_k\}_{k}}\sum_{m=1}^M\log \left|\boldsymbol{\Sigma}_m\right|+\operatorname{tr}\left(\boldsymbol{\Sigma}_m^{-1}\left(\mathbf{y}_m-\bar{\mathbf{y}}_m\right)\left(\mathbf{y}_m-\bar{\mathbf{y}}_m\right)^H\right) \label{obj} \\
& ~~\text{s.t.}~~~~ {a}_{k} \geq 0, ~~~\forall k\in\mathcal{K}_1\cup\mathcal{K}_2\cup\mathcal{K}_3. \label{p1_constraint}
\end{align}
\end{subequations}

\section{The Proposed Mixture of Experts-augmented Deep Unfolding Design}   \label{autoencoder}

In general, rank-one update is a standard way to solve for activity status in the covariance method~\cite{C_ICC_Czl_ad, TWC_Lin2022}. However, since the objective function~\eqref{obj} involves the mean $ \bar{\mathbf{y}}_m$ and the covariance matrix $\boldsymbol{\Sigma}_m$, and they both depend on the activity status $a_k$, rank-one update for estimating $a_k$ is not applicable anymore. Thus, we introduce a projected gradient descent (PGD) approach to solve~\eqref{p}. Furthermore, to rectify the approximation in the likelihood function~\eqref{obj}, the PGD method is unfolded to a neural network so that model inaccuracy could be compensated in a data driven manner. 
\vspace{-3mm}
\subsection{Projected Gradient Dscent Approach}   \label{pgd_fixed }
Noticing that~\eqref{obj} is differentiable with respect to $\{a_k\}_{k\in\mathcal{K}_1\cup\mathcal{K}_2\cup\mathcal{K}_3}$ and the constraint \eqref{p1_constraint} is simple, the PGD approach can provide efficient optimization. In particular, at the $i$-th iteration, the update procedure is written as
\begin{align} \label{eq4}
\hat{a}_{k}^{(i)}=\max \left\{\hat{a}_{k}^{(i-1)}-\eta^{(i)} d_{k}^{(i-1)}, 0\right\},
\end{align}
where $\hat{a}_{k}^{(i)}$ is the estimate of the $k$-th device's activity at the $i$-th iteration, $\eta^{(i)}$ denotes the stepsize, and $d_{k}^{(i-1)}$ is the gradient of \eqref{obj} at $\hat{a}_{k}^{(i-1)}$, which is given by
\begin{equation}\label{grad}
    d_{k}^{(i-1)}=\sum\nolimits^M_{m=1} d_{k,m}^{(i-1)}
\end{equation}
where $d_{k,m}^{(i-1)}$ is given on the top of this page. 

\begin{figure*}
\begin{align} 
d_{k,m}^{(i-1)} &= \operatorname{tr}\left(\Xi_{m,k} \mathbf{s}_{k} \mathbf{s}_{k}^H\left({\boldsymbol{\Sigma}}_m^{(i-1)} \right)^{-1}\right) - \operatorname{tr}\left(\Xi_{m,k} \mathbf{s}_{k} \mathbf{s}_{k}^H\left({\boldsymbol{\Sigma}}_m^{(i-1)} \right)^{-1}\left(\mathbf{y}-\bar{\mathbf{y}}_m^{(i-1)}\right)\left(\mathbf{y}_m-\bar{\mathbf{y}}_m^{(i-1)}\right)^H\left({\boldsymbol{\Sigma}}_m^{(i-1)} \right)^{-1}\right) \nonumber \\
&- \operatorname{tr}\left(\mathbf{s}_{k}   (\mathbf{h}_k^{\text{LoS}})^H \mathbf{\Theta} \mathbf{g}_m^{\text{LoS}}\left(\mathbf{y}_m-\bar{\mathbf{y}}_m^{(i-1)}\right)^H\left({\boldsymbol{\Sigma}}_m^{(i-1)} \right)^{-1}\right) - \operatorname{tr}\left(\left(\mathbf{y}_m-\bar{\mathbf{y}}_m^{(i-1)}\right)\left(\mathbf{s}_{k}   (\mathbf{h}_k^{\text{LoS}})^H \mathbf{\Theta} \mathbf{g}_m^{\text{LoS}}\right)^H\left({\boldsymbol{\Sigma}}_m^{(i-1)} \right)^{-1}\right) \nonumber
\end{align}
\hrulefill
\end{figure*}


\begin{figure} [t]
	\centering 
		 \includegraphics[width=0.48\textwidth]{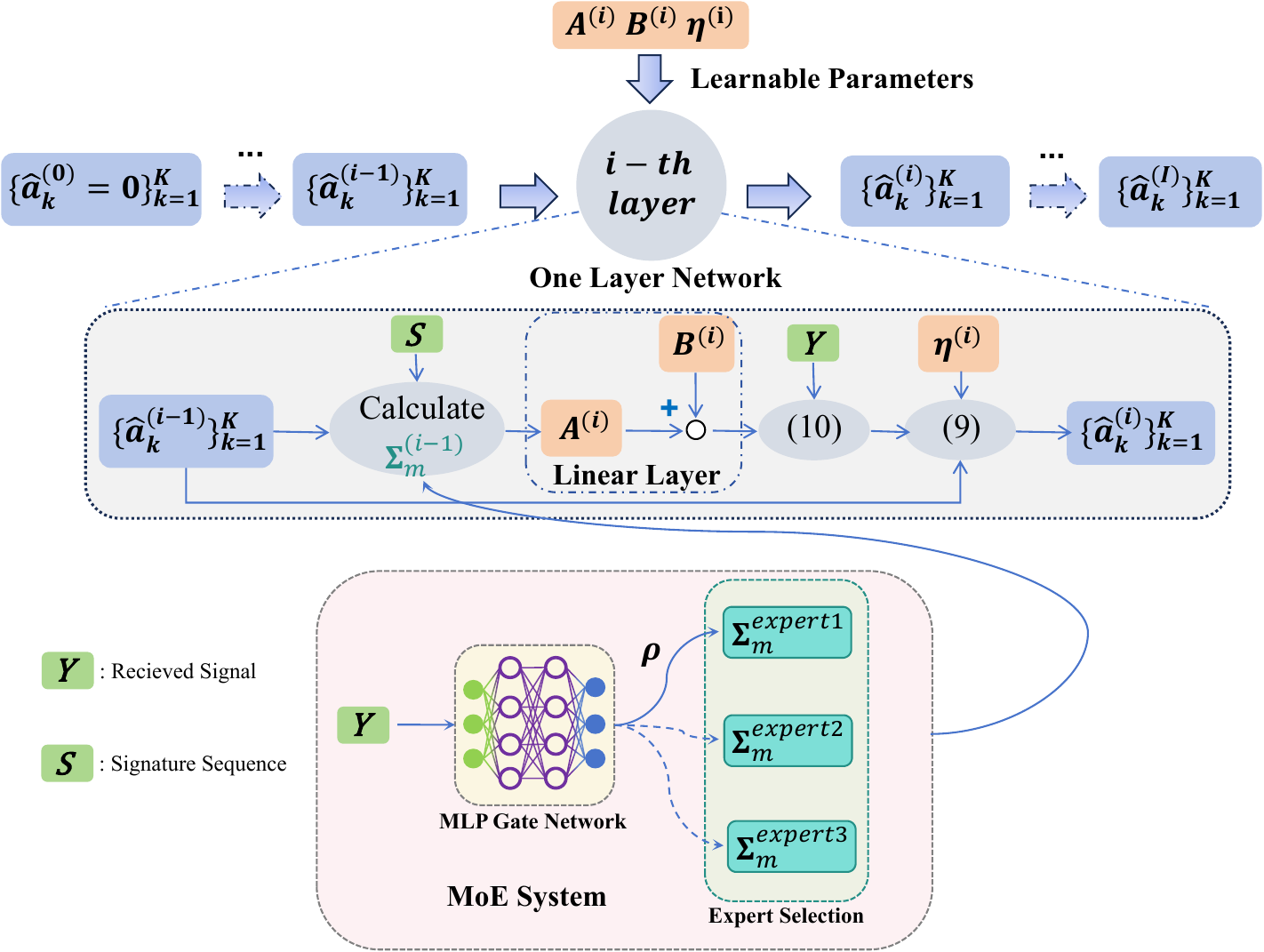}  
	
  	\caption{Architecture of the unfolded network, with details on the feed forward process of the i-th layer and the MoE system. }\vspace{-0.2in}  \label{Fig11}
\end{figure}
\vspace{-2mm}
\subsection{The Deep Unfolded Network}   \label{phase }

Applying deep unfolding to the PGD approach, each iteration of~\eqref{eq4} and~\eqref{grad} is regarded as one layer of the neural network. The key in constructing a deep unfolded network is to introduce trainable parameters in each layer of the neural network. In the PGD algorithm, since the choice of $\eta^{(i)}$ influences the convergence rate and detection performance, we regard it as a trainable parameter. Furthermore, to avoid the matrix inverse in the computation of $d_{k}^{(i-1)}$ and increase the learning capability, we design a linear layer $\mathbf{A}^{(i)}\boldsymbol{\Sigma}_m^{(i-1)} + \mathbf{B}^{(i)}$ with trainable parameters $\mathbf{A}^{(i)}$ and $\mathbf{B}^{(i)}$ to approximate $(\boldsymbol{\Sigma}_m^{(i-1)})^{-1}$. 

The unfolded network consists of $I$ cascaded layers with trainable parameters $\left\{ \mathbf{A}^{(i)},\mathbf{B}^{(i)}, \eta^{(i)}\right\}_{i=1}^{I}$ \textcolor{black}{as shown in~Fig.~\ref{Fig11}}.
Specifically, for the $i$-th layer, the input is $(\mathbf{Y}, \mathbf{S}, \{\hat{a}_{k}^{(i-1)}\}_{k\in\mathcal{K}_1\cup\mathcal{K}_2\cup\mathcal{K}_3})$ and the output is $\{\hat{a}_{k}^{(i)}\}_{k\in\mathcal{K}_1\cup\mathcal{K}_2\cup\mathcal{K}_3}$.

\vspace{-3mm}
\subsection{Mixture of Experts Design}
\label{MoE}

Both the PGD and its deep unfolded version require the computation of $\boldsymbol{\Sigma}_m^{(i-1)}$ in~\eqref{sigma} with the current activity status $\hat{a}_k^{(i-1)}$. However, calculating $\boldsymbol{\Sigma}_m^{(i-1)}$ requires the precise knowledge of which devices belong to which group: $\mathcal{K}_1$, $\mathcal{K}_2$, or $\mathcal{K}_3$. This information is difficult to acquire in practice. One naive way to handle this is to assume all the devices belong to only one of the groups. This corresponds to the covariance matrix being
\begin{align}
\begin{split}
  \boldsymbol{\Sigma}_m^{\text{expert}_1} &= \sum_{k\in\mathcal{K}_1\cup\mathcal{K}_2\cup\mathcal{K}_3} {a_k} \Xi_{m,k} \mathbf{s}_{k} \mathbf{s}_{k}^H+\sigma_w^2 \mathbf{I}_M,\\
  \boldsymbol{\Sigma}_m^{\text{expert}_2} &= \sum_{k\in\mathcal{K}_1\cup\mathcal{K}_2\cup\mathcal{K}_3} \frac{a_k\mathbf{s}_{k} \mathbf{s}_{k}^H}{1+\kappa_{\text{U}}}+\sigma_w^2 \mathbf{I}_M,\\
  \boldsymbol{\Sigma}_m^{\text{expert}_3} &= \sum_{k\in\mathcal{K}_1\cup\mathcal{K}_2\cup\mathcal{K}_3} {a_k}  \mathbf{s}_{k} \mathbf{s}_{k}^H+\sigma_w^2 \mathbf{I}_M. \nonumber
\end{split}
\end{align}

However, how do we know which covariance matrix we should use in practice? To address this challenge, we propose a MoE design. However, one of the major challenges in constructing a MoE architecture lies in designing the gate network to process input information and assigning the most appropriate expert for subsequent computation. To this end, we propose a MLP as the gating network. Upon feeding the received signal matrix ${\mathbf{Y}}$ into the gating network, it outputs the proportions of three types of devices. By detecting the device type with the highest proportion, we select the corresponding expert to compute the covariance matrix. The specific structure of the MoE system and its integration with the unfolded network are illustrated in Fig.~\ref{Fig11}. By introducing this MoE-module,  it is not necessary to acquire the channel type information between each device and the BS anymore.

The MoE framework aligns with the mixture-of-experts theory~\cite{chen2022nips,fedus2022jmlr}, where the gating network learns to partition the input space into regions dominated by specific experts. By minimizing the KL divergence loss:
\begin{equation}
\mathcal{L} = \sum\nolimits_{i=1}^3 \rho_i^{\text{true}} \log \left({\rho_i^{\text{true}}}/{\rho_i}\right),
\end{equation}
the gating network asymptotically approximates the true device group proportions,
\begin{equation}
\rho_i^{\text{true}} = {|\mathcal{K}_i|}/{\sum\nolimits_{i=1}^3 |\mathcal{K}_i|}, \quad i = 1,2,3.
\end{equation}
This ensures that the selected expert matches the dominant channel type, thereby adapting to mixed fading conditions. As analyzed in~\cite{chen2022nips}, this nonlinear top-1 expert selection MoE performs well on classification problems.

The MoE-augmented deep unfolding network is trained end-to-end using incremental training~\cite{twc_LIN}. The gating network's stability is ensured by the convexity of the KL divergence loss, which guarantees that gradient-based updates converge to a local minimum~\cite{fedus2022jmlr}. The overall network convergence is further guaranteed by the convergence properties of the PGD algorithm~\cite{twc_LIN}.


The computational complexity of the deep unfolding network without the MoE module is $\mathcal{O}(I \cdot (M^3 + LM))$. And with the MoE module, the computational complexity becomes $\mathcal{O}(LM + I \cdot (M^3 + LM))$. Given that $LM \ll I \cdot M^3$ (with $L = 20$, $M = 32$ and $I = 4$ in the simulation), the impact of the MoE module on the total complexity is negligible.

\vspace{-3mm}
\section{Simulation Results} \label{sim}

In this section, we compare the performance of the proposed MoE-augmented deep unfolding approach with two deep unfolding baselines~\cite{sun2024baseline1,zou2024baseline2}, traditional coordinate descent (CD) algorithm (assume all channels are zero-mean Gaussian), the derived PGD algorithm and a transformer-based detector~\cite{LY_transformer}. After obtaining the estimate $\hat{a}_{k}$ for each device, the estimated activity $\hat{b}_{k}=1$ if $\hat{a}_{k} \geq a^{t h}$ (otherwise, $\hat{b}_{k}=0$),
where $a^{th}$ is a threshold that controls the trade-off between the probabilities of false alarm $\mathrm{PF}=\frac{\sum_{k=1}^K \hat{b}_{k}\left(1-{b}_{k}\right)}{\sum_{k=1}^K\left(1-{b}_{k}\right)}$ and missed detection $\mathrm{PM}=1-\frac{\sum_{k=1}^K {b}_{k} \hat{b}_{k}}{\sum_{k=1}^K{b}_{k}}$~\cite{C_ICC_Czl_ad}.

Under a three dimensional Cartesian coordinate system, we set the locations of the BS and the IRS at $(0,0,10)$ and $(5,50,10)$ in meter, respectively. In addition, devices from $\mathcal{K}_1$ are randomly and uniformly located in a circular area of radius $40$ m around the center $(200,0,0)$. The devices in $\mathcal{K}_2$ and $\mathcal{K}_3$ are located in the left half circular area with radius $40$m around the center $(0,120,0)$. The large-scale coefficient of each device $\beta_k$ is generated according to $\beta_k = -60 - 22\log_{10}(\lambda_{k}\lambda_{0})$ in dB, where $\lambda_{k}$ denotes the distance between device~$k$ and the IRS, and $\lambda_{0}$ denotes the distance between the IRS and the BS. The transmit power of each device is set as $p_k = 23$ dBm. The BS is equiped with $32$ antennas and the IRS is with $40$ reflecting elements. The number of potential devices is $K = 100$, with the activity probability being 0.2. The length of signature sequence $\mathbf{s}_k$ is set at $L = 20$.


To train an MLP-based gate network, the received signal matrix $\mathbf{Y} \in \mathbb{C}^{L \times M}$ is first preprocessed by concatenating real and imaginary components into a real-valued vector $\mathbf{y}_{\text{real}} = [\text{Re}(\text{vec}(\mathbf{Y})); \text{Im}(\text{vec}(\mathbf{Y}))] \in \mathbb{R}^{2LM}$. The MLP for MoE consists of three fully-connected layers: an input layer ($2LM \to 512$) with ReLU activation, a hidden layer ($512 \to 128$) with tanh activation, and an output layer ($128 \to 3$) with softmax normalization to produce $\rho_1$, $\rho_2$, $\rho_3$ as the estimate of user group proportions. For the deep unfolding network, we use incremental training~\cite{twc_LIN}, and it is found that the network with more than 4 layers only leads to a small performance improvement. Therefore, we choose $I=4$. 
\begin{figure} [t]
	\centering
 	{\label{fig:moe} 
		\includegraphics[width=0.4\textwidth]{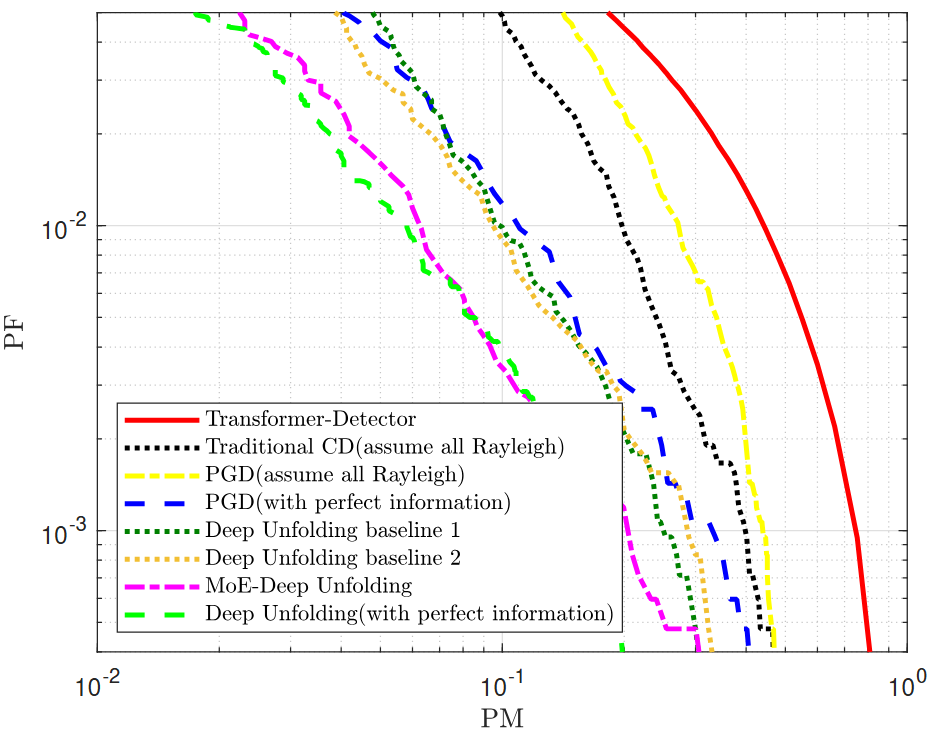}}
	  
  	\caption{Performance comparison in terms of PM and PF.}\vspace{-0.2in}  \label{fig:moe}
\end{figure}
\begin{figure} [t]
	\centering
 	{\label{fig:1} 
		\includegraphics[width=0.4\textwidth]{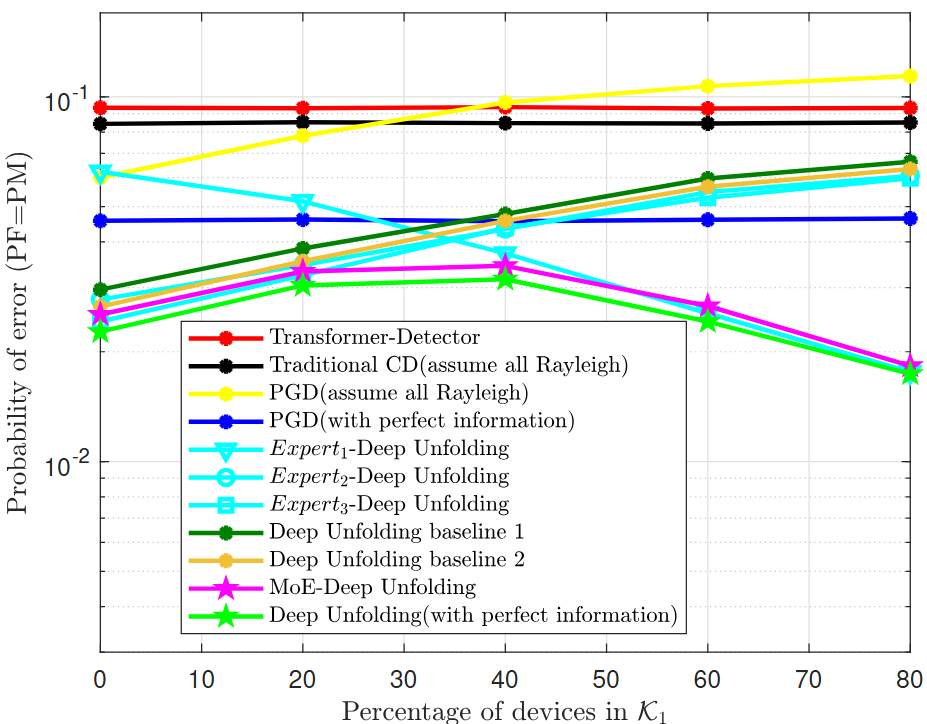}}
	  
  	\caption{PF = PM versus percentage of devices in $\mathcal{K}_1$.}\vspace{-0.2in}  \label{fig:2}
\end{figure}

First, we present the PM-PF curves of various methods under the setting of: $\kappa_\text{U}$ = $\kappa_\text{R}$ = $\kappa_\text{B}$ = 10 dB, $\sigma_w^2$ = -95 dBm. Among 100 devices, 40\% of them are from $\mathcal{K}_1$ and the proportions of devices in $\mathcal{K}_2$ and $\mathcal{K}_3$ are both 30\%. In Fig.~\ref{fig:moe}, it can be observed that the proposed deep unfolding design significantly outperforms both the optimization-based PGD and CD methods as the deep unfolded network could compensate the inaccurate modeling. Furthermore, the deep unfolding method is superior than the transformer based detector, since deep unfolding utilizes the mathematical principles as implicit training guidelines, making it easier to achieve better detection performance. While the original deep unfolding method still needs specific channel type information between devices and the BS, the MoE-aided deep unfolding network, which does not need the knowledge of which device belongs to which group, only suffers from a small degradation compared to the deep unfolded network with perfect channel type information. In contrast, if no such information is available for the PGD algorithm, it has a significant degradation compared to the PGD with perfect information. When compared with the two unfolding baselines, the proposed unfolding network performs better as it is designed specifically for mixed channel types scenarios while the baselines are designed for one type of channel only. 



Next, Fig.~\ref{fig:2} illustrates the flexibility and generalization ability of the proposed MoE-augmented deep unfolding approach. We compare the error rates of various methods with the proportion of devices in $\mathcal{K}_1$ varies from 0\% to 80\%. The remaining devices is equally divided between $\mathcal{K}_2$ and $\mathcal{K}_3$. We can observe that the deep unfolding network that uses a single fixed expert only performs well if the network setting matches the expert's prior knowledge. However, with the MoE system enabled, the gate network can learn which type of device has the highest proportion and select the best expert accordingly. This property leads to the MoE-based deep unfolding network performing close to the deep unfolding network with perfect devices' grouping information. For the two deep unfolding baselines, they perform similar to the proposed unfolded PGD method with the covariance matrix fixed as expert 3. This is not surprising as the two baselines also assume all users experience Rayleigh fading.

\begin{figure}[t]
\centering
\includegraphics[width=0.33\textwidth]{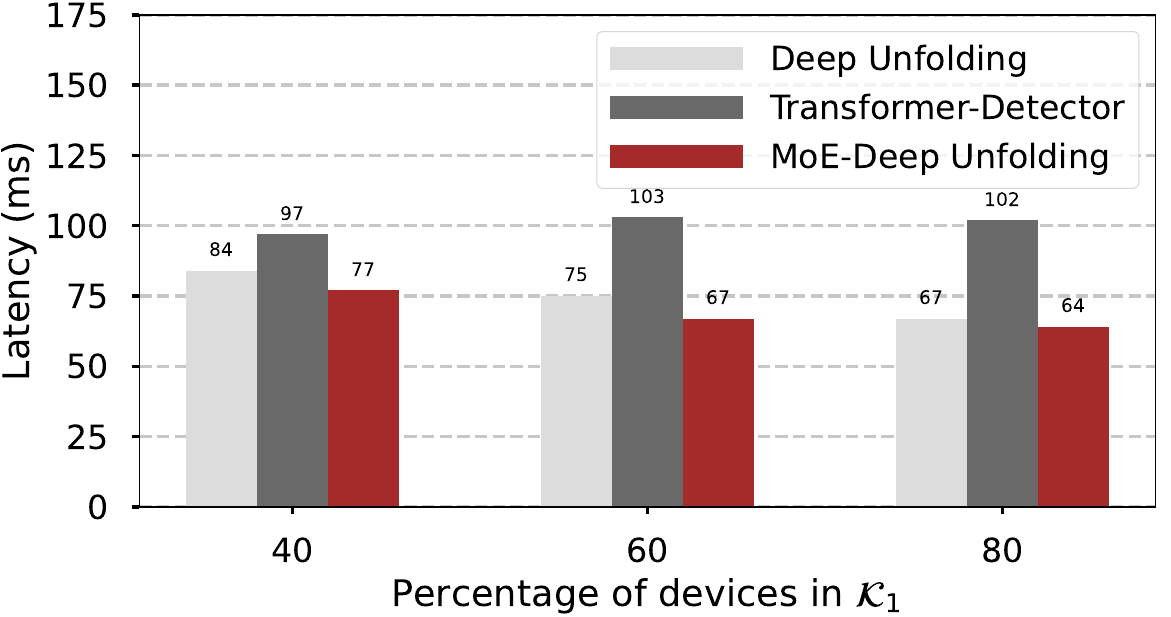}
\caption{Latency versus percentage of devices in $\mathcal{K}_1$}\vspace{-0.2in}
\label{fig3}
\end{figure}

Finally, we present the computation times of different deep learning methods. As shown in Fig.~\ref{fig3}, the MoE-augmented deep unfolding approach experiences the shortest inference time among three deep learning methods. Comparing to the deep unfolding approach with perfect channel type information, the proposed MoE approach, although involves an additional MLP, exhibits even faster computation time since the covariance computation is simpler by focusing on the dominant channel type in the network. Meanwhile, transformer-based detector experiences the slowest inference time due to its generic structure.

\vspace{-3mm}
\section{Conclusions}
\vspace{-2mm}
In this paper, we proposed a model-driven deep learning approach for activity detection in IRS-aided systems with mixed types of channel fading. Specifically, we derived an approximated covariance-based formulation and introduced a deep unfolding network based on the projected gradient method. By further incorporating a MoE network, the proposed method does not require prior knowledge of which device belongs to which channel type. Simulation results showed that the MoE-augmented model-driven deep unfolding method achieves better detection performance than traditional covariance-based methods and black-box neural network designs. Furthermore, the performance loss from not knowing the fading channel type of each device is minimal compared to the unfolded network with perfect channel type information.

\vspace{-3mm}
\bibliographystyle{IEEEtran}
\bibliography{ref}

\begin{thebibliography}{10}
\providecommand{\url}[1]{#1}
\csname url@samestyle\endcsname
\providecommand{\newblock}{\relax}
\providecommand{\bibinfo}[2]{#2}
\providecommand{\BIBentrySTDinterwordspacing}{\spaceskip=0pt\relax}
\providecommand{\BIBentryALTinterwordstretchfactor}{4}
\providecommand{\BIBentryALTinterwordspacing}{\spaceskip=\fontdimen2\font plus
\BIBentryALTinterwordstretchfactor\fontdimen3\font minus
  \fontdimen4\font\relax}
\providecommand{\BIBforeignlanguage}[2]{{%
\expandafter\ifx\csname l@#1\endcsname\relax
\typeout{** WARNING: IEEEtran.bst: No hyphenation pattern has been}%
\typeout{** loaded for the language `#1'. Using the pattern for}%
\typeout{** the default language instead.}%
\else
\language=\csname l@#1\endcsname
\fi
#2}}
\providecommand{\BIBdecl}{\relax}
\BIBdecl

\bibitem{isacqf}
X.~Luo, Q.~Lin, R.~Zhang, H.-H. Chen, X.~Wang, and M.~Huang, ``Isac – a
  survey on its layered architecture, technologies, standardizations,
  prototypes and testbeds,'' \emph{IEEE Communications Surveys \& Tutorials},
  pp. 1--1, 2025.

\bibitem{Liuliang_TSP_FUTURE}
L.~Liu, E.~G. Larsson, W.~Yu, P.~Popovski, \v{C}. Stefanovi\'{c}, and
  E.~de~Carvalho, ``Sparse signal processing for grant-free massive
  connectivity: A future paradigm for random access protocols in the internet
  of things,'' \emph{IEEE Signal Process. Mag.}, vol.~35, no.~5, pp. 88--99,
  Sep. 2018.

\bibitem{tsp_hao}
H.~Zhang, Q.~Lin, Y.~Li, L.~Cheng, and Y.-C. Wu, ``Activity detection for
  massive connectivity in cell-free networks with unknown large-scale fading,
  channel statistics, noise variance, and activity probability: A bayesian
  approach,'' \emph{IEEE Trans. Signal Process.}, vol.~72, pp. 942--957, 2024.

\bibitem{liyang_ad}
Y.~Li, Q.~Lin, Y.-F. Liu, B.~Ai, and Y.-C. Wu, ``Asynchronous activity
  detection for cell-free massive mimo: From centralized to distributed
  algorithms,'' \emph{IEEE Trans. Wireless Commun.}, vol.~22, no.~4, pp.
  2477--2492, 2023.

\bibitem{tspll}
L.~Liu and W.~Yu, ``Massive connectivity with massive \protect{MIMO}—part
  \protect{I}: Device activity detection and channel estimation,'' \emph{IEEE
  Trans. Signal Process.}, vol.~66, no.~11, pp. 2933--2946, Jun. 2018.

\bibitem{haghighatshoar2018}
S.~Haghighatshoar, P.~Jung, and G.~Caire, ``Improved scaling law for activity
  detection in massive \protect{MIMO} systems,'' in \emph{IEEE International
  Symposium on Information Theory (ISIT)}, 2018.

\bibitem{C_ICC_Czl_ad}
Z.~Chen, F.~Sohrabi, Y.-F. Liu, and W.~Yu, ``Covariance based joint activity
  and data detection for massive random access with massive \protect{MIMO},''
  in \emph{IEEE Int. Conf. Commun. (ICC)}, 2019.

\bibitem{TWC_Lin2022}
Q.~Lin, Y.~Li, and Y.-C. Wu, ``Sparsity constrained joint activity and data
  detection for massive access: {A} difference-of-norms penalty framework,''
  \emph{IEEE Trans. Wireless Commun.}, vol.~22, no.~3, pp. 1480--1494, Mar.
  2023.

\bibitem{tit_c}
Z.~Chen, F.~Sohrabi, Y.-F. Liu, and W.~Yu, ``Phase transition analysis for
  covariance-based massive random access with massive {MIMO},'' \emph{IEEE
  Trans. Inf. Theory}, vol.~68, no.~3, pp. 1696--1715, Mar. 2022.

\bibitem{J_TWC_shaoxiaodan_risad}
X.~Shao, L.~Cheng, X.~Chen, C.~Huang, and D.~W.~K. Ng, ``Reconfigurable
  intelligent surface-aided {6G} massive access: Coupled tensor modeling and
  sparse {Bayesian} learning,'' \emph{IEEE Trans. Wireless Commun.}, vol.~21,
  no.~12, pp. 10\,145--10\,161, Dec. 2022.

\bibitem{J_TSP_shiyuanming_risad}
S.~Xia, Y.~Shi, Y.~Zhou, and X.~Yuan, ``Reconfigurable intelligent surface for
  massive connectivity: Joint activity detection and channel estimation,''
  \emph{IEEE Trans. Signal Process.}, vol.~69, pp. 5693--5707, Oct. 2021.

\bibitem{Sun_TSP}
H.~Sun, X.~Chen, Q.~Shi, M.~Hong, X.~Fu, and N.~D. Sidiropoulos, ``Learning to
  optimize: Training deep neural networks for interference management,''
  \emph{IEEE Trans. Signal Process.}, vol.~66, no.~20, pp. 5438--5453, Aug.
  2018.

\bibitem{lei}
J.~Lei, Y.~Li, L.-Y. Yung, Y.~Leng, Q.~Lin, and Y.-C. Wu, ``Understanding
  complex-valued transformer for modulation recognition,'' \emph{IEEE Wireless
  Commun. Letters}, vol.~13, no.~12, pp. 3523--3527, 2024.

\bibitem{Du_MoE}
H.~Du, G.~Liu, Y.~Lin, D.~Niyato, J.~Kang, Z.~Xiong, and D.~I. Kim, ``Mixture
  of experts for intelligent networks: A large language model-enabled
  approach,'' in \emph{2024 International Wireless Communications and Mobile
  Computing (IWCMC)}, 2024.

\bibitem{wc_du}
J.~Wang, H.~Du, D.~Niyato, J.~Kang, Z.~Xiong, D.~I. Kim, and K.~B. Letaief,
  ``Toward scalable generative ai via mixture of experts in mobile edge
  networks,'' \emph{IEEE Wireless Commun.}, vol.~32, no.~1, pp. 142--149, 2025.

\bibitem{twc_LIN}
Q.~Lin, Y.~Li, Y.-C. Wu, and R.~Zhang, ``Intelligent reflecting surface aided
  activity detection for massive access: Performance analysis and learning
  approach,'' \emph{IEEE Trans. Wireless Commun.}, vol.~23, no.~11, pp.
  16\,935--16\,949, 2024.

\bibitem{chen2022nips}
Z.~Chen, Y.~Deng, Y.~Wu, Q.~Gu, and Y.~Li, ``Towards understanding mixture of
  experts in deep learning,'' in \emph{Proc. Int. Conf. Neural Inform. Process.
  Syst. (NeurIPS)}, 2022.

\bibitem{fedus2022jmlr}
W.~Fedus, B.~Zoph, and N.~Shazeer, ``Switch transformers: scaling to trillion
  parameter models with simple and efficient sparsity,'' \emph{J. Mach. Learn.
  Res. (JMLR)}, vol.~23, no.~1, Jan. 2022.

\bibitem{sun2024baseline1}
G.~Sun, W.~Wang, W.~Xu, and C.~Studer, ``Deep-unfolded joint activity and data
  detection for grant-free transmission in cell-free systems,'' in \emph{2024
  19th International Symposium on Wireless Communication Systems (ISWCS)},
  2024, pp. 1--5.

\bibitem{zou2024baseline2}
Y.~Zou, Y.~Zhou, X.~Chen, and Y.~C. Eldar, ``Proximal gradient-based unfolding
  for massive random access in iot networks,'' \emph{IEEE Trans. Wireless
  Commun.}, vol.~23, no.~10, pp. 14\,530--14\,545, 2024.

\bibitem{LY_transformer}
Y.~Li, Z.~Chen, Y.~Wang, C.~Yang, B.~Ai, and Y.-C. Wu, ``Heterogeneous
  transformer: A scale adaptable neural network architecture for device
  activity detection,'' \emph{IEEE Trans. Wireless Commun.}, vol.~22, no.~5,
  pp. 3432--3446, May 2023.

\end{thebibliography}
\vfill

\end{document}